# The meaning of prompts and the prompts of meaning: Semiotic reflections and modelling

## Abstract

This paper explores prompts and prompting in large language models (LLMs) as dynamic semiotic phenomena, drawing on Peirce's triadic model of signs, his nine sign types, and the Dynacom model of communication. The aim is to reconceptualize prompting not as a technical input mechanism but as a communicative and epistemic act involving an iterative process of sign formation, interpretation, and refinement. The theoretical foundation rests on Peirce's semiotics, particularly the interplay between representamen, object, and interpretant, and the typological richness of signs: qualisign, sinsign, legisign; icon, index, symbol; rheme, dicent, argument - alongside the interpretant triad captured in the Dynacom model . Analytically, the paper positions the LLM as a semiotic resource that generates interpretants in response to user prompts, thereby participating in meaning-making within shared universes of discourse. The findings suggest that prompting is a semiotic and communicative process that redefines how knowledge is organized, searched, interpreted, and co-constructed in digital environments. This perspective invites a reimagining of the theoretical and methodological foundations of knowledge organization and information seeking in the age of computational semiosis

**Martin Thellefsen, Amalia Nurma Dewi & Bent Sørensen**

## 1. Introduction

Recent advancements in artificial intelligence (AI), particularly the widespread acceptance of large language models (LLMs), have become increasingly prominent in our daily life as well as in academia. This technology has had a significant impact on how we think about AI and LLMs and altered the scene of human-computer interaction, establishing the "prompt" as a pivotal element in modern digital communication. This development represents more than a mere technological advancement; it constitutes a fundamental shift in how humans engage with computational systems in order to access, manipulate, and generate knowledge. Although the architecture of AI is built on sophisticated neural networks and statistical pattern recognition, the interaction follows a path that is inherently communicative, in that the interaction mimic dialogic negotiations of meaning between a human actor and the LLM system.

The emergence of *prompt engineering* as a distinct practice reflects this new communicative complexity. Practitioners have e.g., developed increasingly advanced techniques for crafting prompts that elicit desired responses from LLMs, that range from simple instructions to complex multi-step reasoning chains and few-shot learning examples (Wei et al., 2022; Zhou et al., 2023). However, the field remains largely empirical and atheoretical and is characterized by trial-and-error methodologies, heuristic guidelines, and community-driven best practices rather than a theoretical foundation (Collins et al., 2021; Sahoo et al., 2024; Wang et al., 2023). The absence of a theoretically based communicative framework creates several critical gaps in our understanding: it obscures the underlying dynamics of meaning negotiation between human and machine; it fails to account for the interpretive processes that govern successful prompt-response cycles; and it provides no systematic basis for predicting or optimizing interaction outcomes (Binz & Schulz, 2023; Qin et al., 2023).

This theoretical lacuna is particularly problematic given the increasing integration of LLMs into knowledge work, educational contexts, and information systems. Without a robust conceptual framework, practitioners lack methods for understanding why certain prompts succeed while others fail. In this paper we argue that the theory of semiotics, specifically the pragmatic semiotics of Charles Sanders Peirce (1839-1914), provides a robust and theoretically grounded framework for understanding prompt-based interactions with LLMs. Peirce's semiotics provides a systematic method for analyzing sign relations, accounts for the dynamic and interpretive nature of meaning-making and offers precise analytical categories that can be applied in computational contexts understood as involving communicative processes  (Bergman, 2018; Tanaka-Ishii, 2010).

We propose that the prompt functions not merely as a computational instruction or query, but as a complex sign – what Peirce termed a Representamen – that initiates a dynamic process of sign-action, or semiosis. This reconceptualization transforms the prompt from a simple input mechanism into a complex communicative artifact that embodies the user's intentions, references specific objects or concepts, and operates within particular interpretive contexts. The LLM's response, in this framework, thus functions as the sign's Interpretant – not merely an output, but as an active interpretation that mediates between the prompt-sign and its intended meaning. The user's intention, along with the real-world referents and conceptual domains that the prompt addresses, constitute the sign's Object, completing Peirce's fundamental triadic relation. Viewing prompt-LLM interaction through this semiotic lens consequently transforms our understanding of the process from a mechanistic input-output operation into a (potentially) rich, interpretive, and inherently meaningful exchange.

The first objective of this analysis, then, is to systematically deconstruct the semiotic processes underlying prompt-based interactions with LLMs. By applying Peirce's systematic classification of signs according to their internal structure and relational properties we thus propose an analytical framework for prompts as (combinations) of different sign types which concerns three (inter-dependent) perspectives. That is, understanding the prompt as 1) a sign relation in itself (via quali-, sin-, and legisign), 2) representing an object (qua icon, index, and symbol), and 3) representing an object for an interpretant (qua rheme, dicent and argument).  We call these three perspectives, inspired by Liszka (1996), the presentative, representative and interpretative dimensions of the prompt as sign.

Furthermore, we move from the formal analysis of the prompt as sign and sign types to prompting explicitly understood as communication; that is, we turn to the Peircean inspired "Dynacom model" (Thellefsen, Sørensen & Thellefsen, 2011). The Dynacom conceptualizes communication as a process involving the transmission and transformation of signs across interpretants. In this model, prompting is not merely the issuance of a sign but the initiation of a communicative act that unfolds through interpretation, response, and feedback.

The paper, therefore, is organized in three main analytical phases. Firstly, we outline the Peircean semiotic framework and establish the theoretical foundations necessary for our analysis and it's its applicability to computational contexts. Secondly, we apply this framework to analyze the prompt as a (quasi) static sign, examining its triadic structure and classificatory properties according to Peirce's systematic typology. Thirdly, we extend our analysis to examine the prompt-response interaction as a chain of semiosis, tracing how meaning unfolds through potential successive interpretative acts – qua different types of interpretants relative to collateral experience and a universe of discourse – all captured in the model of communication, the “Dynacom”.

Finally, and more briefly, we conclude the paper by discussing the broader implications of this semiotic perspective for information science and knowledge organization, positioning the LLM as a novel type of semiotic resource that challenges the traditional boundaries between human and machine interpretation. The analysis contributes to the emerging discussion about the epistemological status of AI-generated knowledge, the nature of human-machine communication, and the theoretical foundations needed to understand increasingly sophisticated forms of computational semiosis – such as prompts and prompting.

## 2. Theoretical framework: A Peircean lens for prompts and prompting

### 2.1. The prompt as sign phenomenon

The analysis of the “prompt-response” paradigm in large language models requires a theoretical framework that is able to account for the unique nature of this phenomenon of meaning-making and communication. Unlike traditional human-computer interaction, prompting is not a simple transmission of information. Instead, it can be considered a (quasi) negotiation of meaning between two distinct cognitive architectures: Human intentionality and machine pattern recognition. In the context of prompting, the fundamental semiotic components can with Peirce be defined and operationalized triadically as follows:

The Representamen (sign proper)**:** This constitutes the material (perceptual), intelligible form of the sign itself - the actual string of text, code, image, or multimodal input that the user submits to the LLM. It is the physical or digital artifact that initiates the semiotic process and serves as the immediate interface between human intention and machine processing. The Representamen encompasses not only the semantic content of the prompt but also its syntactic structure, formatting, contextual positioning, and any embedded instructions or examples. Importantly, the Representamen is not merely a neutral container for information but actively shapes the interpretive process through its specific material and structural properties.

The object**:** This represents that to which the sign refers, encompassing both the intended referent and the broader reality that the sign attempts to capture or manipulate (Peirce, CP 2.232). Following Peirce’s distinction, we differentiate between the immediate object and the dynamic object (Peirce, CP 4.536). The immediate object is the object as it is represented within and by the sign itself - in prompting contexts, this corresponds to the user’s intention as it is captured and encoded within the prompt’s textual or multimodal structure. It includes the specific aspects of the user’s goal that are made explicit in the prompt, the particular framing or perspective adopted, and the constraints or parameters specified. The dynamic object, by contrast, is the object as it has its being independently of any particular sign – the broader reality, concept, dataset, or domain of knowledge that the user ultimately hopes to access, manipulate, or understand. This might include the full complexity of a scientific domain, the nuanced cultural context of a literary work, or the comprehensive structure of a database that extends far beyond what can be captured in any single prompt.

The interpretant**:** This element represents the effect of the sign on an interpreter – the meaning of the sign as it is realized through interpretive action (Peirce, CP 1.339). In the primary instance of prompt-response interaction, the LLM functions as the interpreter, and the interpretant manifests as the model’s processing of the prompt and its eventual textual or multimodal output. (It is crucial to note that this interpretation is a function of sophisticated statistical pattern matching, not of conscious understanding; yet, this does not prevent the LLM from participating meaningfully in the semiotic process). However, the Interpretant is not simply equivalent to the model's response; it encompasses the entire interpretive process, including the activation of relevant parameters, the retrieval and synthesis of training data patterns, and the generation strategies employed. Crucially, this output then becomes a new representamen for the human user, who must interpret the model's response in relation to their original intention, thus creating a continuous chain of semiosis that can extend through multiple conversational turns (Peirce, CP 2.303; Eco, 1976). This recursive structure captures the iterative nature of effective prompting, where users refine their prompts based on the model's response, and models generate responses that build upon previous exchanges.

**2.2 Prompts as types of signs**

Whether in natural language interfaces, programming environments, or artistic creation, prompts function as signs that elicit representation, interpretation and (communicative) action. To understand their nature more precisely, we first turn to Peirce's systematic typology of signs. Based on his tri-relative understanding of the sign – involving representamen, object, interpretant – signs can be analyzed from three (inter-dependent) dimensions; that is, the representamen as such, the relation between the representamen of the object and the relation between the representamen and the interpretant. And according to Peirce these three dimensions yield three typologies of signs, namely, qualisign, sinsign, legisign (first sign typology); icon, index, symbol (second sign typology); rheme, dicent, argument (third sign typology). Briefly formulated the nine signs of the three sign typologies can be described as follows (Peirce, 1998: 290-294):

The first typology concerns the ontological mode of the representamen in itself. A qualisign is a representamen insofar as it is a pure quality, a possibility of presentation that is not yet actualized; a sinsign is a representamen as a singular, concrete occurrence, an actual instantiation of a sign-event; finally, a legisign is a representamen as a general rule or law, a conventional or habitual pattern that governs the production and recognition of signs across instances. This typology addresses, then, how representamina present themselves to perception and cognition, whether as potential, actual, or general phenomena.

The second typology pertains to the relation between the representamen and its object, and thus constitutes the representative dimension of the sign. An icon signifies by virtue of a relation of similarity to its object. An index signifies through a factual or existential connection to its object, often involving contiguity or causality. A symbol signifies through convention, habit or learned association, relying on a shared system of meaning.

Finally, the third typology addresses the relation between the representamen and the interpretant, forming the interpretative dimension of the sign. A rheme, then, is a sign that presents a possibility or potential meaning, without asserting or reasoning. A dicent is a sign that makes a claim or assertion, presenting a proposition that can be evaluated as true or false. Whereas an argument is a sign that conveys a logical relation, guiding the interpretant through a process of inference or reasoning. Thus, this typology reveals how signs function in the process of interpretation, shaping the cognitive and inferential engagement of the interpreter.

The table 1 below gives an overview over relations between the nine sign types and their central characteristics:

| 1st sign typology | 2nd sign typology | 3rd sign typology |
|---|---|---|
| *Qualisign*<br>in itself, the sign is of the nature of appearance. | *Icon*<br>a sign which refers to the object merely by virtue of characters of its own. | *Rheme*<br>a sign which, for its interpretant, is a sign of possibility. |
| *Sinsign*<br>in itself, the sign is of the nature of an individual object or fact. | *Index*<br>a sign which refers to the object by virtue of some existential relation. | *Dicent*<br>a sign which, for its Interpretant, is a sign of actual existence. |
| *Legisign*<br>in itself, the sign is of the nature of a general type. | *Symbol*<br>a sign which refers to the object by virtue of some kind of convention. | *Argument*<br>a sign which, for its Interpretant, is a sign of law. |

Table 1: Peirce's nine signs organized in the three typologies (after Queiros, 2012: 57)

Peirce's three typologies of signs offer, we believe, a nuanced typology that can enrich our understanding of prompts. Inspired by Liszka (1996) we argue that, these typologies correspond to three analytical dimensions, namely, the presentative, representative, and interpretative dimensions of the prompt, respectively. Or formulated differently, the theoretical and methodological value of this semiotic framework lies in its capacity to reveal the prompt as a layered, relational, and interpretative sign. It moves us beyond surface-level analysis and into a deeper understanding of how prompts are potentially meaning-making. Let us therefore below address each of these three semiotic dimensions of the prompt one by one.

#### 2.2.1. Prompts and the first sign typology (the presentive dimension)

Together, the qualisign, sinsign and legisign allow us to begin analyzing the prompt's formal and experiential presentation, revealing how it is constructed and how it (potentially) enters into communicative circulation concerning human-machine interaction.

The *qualisign*, as Peirce defines it, refers to a quality that functions as a sign. It is not an actual sign until it is embodied in a concrete instance, but it nonetheless carries semiotic potential. In the context of LLMs, the qualisign dimension of a prompt encompasses its affective, stylistic, and tonal characteristics – those features that shape the expressive force of the prompt and influence the model's generative stance. Although the model does not experience affect in the human sense, it is trained to recognize and respond to stylistic cues embedded in language. These cues guide the model's output, shaping its rhetorical posture and lexical choices.

Consider the following pair of prompts:

- "Describe the feeling of walking through a foggy forest at dawn."
- "List the environmental factors contributing to fog formation in temperate forests."

Both prompts refer to fog and forests, yet the first evokes a sensory, atmospheric experience. It invites the model to adopt a poetic or narrative register, perhaps drawing on literary tropes or emotional language. The second prompt, by contrast, signals a scientific and factual tone, encouraging the model to produce an analytical response grounded in meteorological terminology.

Another example might be:

- "Write a love letter to the moon."
- "Summarize the gravitational effects of the moon on Earth's tides."

Here, the first prompt is rich in qualisign potential – it suggests metaphor, emotion, and imaginative language (Sørensen & Thellefsen 2006; Sørensen, Thellefsen & Moth 2007). The second is stripped of affective resonance and oriented toward technical explanation. The model, trained on vast corpora of text, can detect these stylistic cues and adjust its output accordingly, even though the qualisign itself remains a latent possibility until instantiated. These examples illustrate how the qualisign operates as a latent possibility – an expressive potential that shapes the model's interpretative orientation. The prompt's tone, mood, and stylistic register are not signs in themselves until they are instantiated, but they nonetheless exert a formative influence on the model's response.

The *sinsign*, by contrast, is the prompt as an actualized event. It is the specific linguistic or symbolic input that the user provides to the model at a given moment. The sinsign is situated in time and space, addressed to a particular instantiation of the model, and exists only in its moment of use. It is the empirical instantiation of the prompt – the site where the qualisign becomes embodied and where the legisign is enacted. Each prompt entered into an LLM interface is a sinsign: a unique occurrence that triggers a computational response. The sinsign is where semiotic potential becomes actualized, and where the model engages with the prompt as a discrete communicative act.

For example:

- A user types: "Generate a haiku about winter solitude."

This is a sinsign. It is a discrete, time-bound event that activates the model's poetic generation capabilities. The prompt draws upon the qualisign of poetic minimalism and seasonal imagery, and it also relies on legisign conventions (such as the structure of a haiku) to guide the model's output.

Another sinsign might be:

- "Correct the grammar in the following sentence: 'He go to school every day.'"

This prompt is an actualized request for grammatical correction. It exists only in its moment of use, yet it invokes broader conventions of English grammar and the expectation that the model can perform linguistic editing. The sinsign is where the abstract qualities of the qualisign and the general rules of the legisign converge in a specific communicative act.

Even more contextually rich examples include:

- "You're a medieval scribe. Translate this modern phrase into Old English: 'The king is wise.'"
- "Pretend you're a sarcastic teenager. React to the sentence: 'Homework is fun.'"

Each of these prompts is a sinsign that activates a particular persona or stylistic mode within the model. They are singular events that rely on the model's ability to simulate voice, tone, and historical or cultural context. Or perhaps more precisely formulated: These prompts show how the sinsign functions as the prompt in its actualized form – a discrete, empirical sign that exists only in its moment of use, yet draws upon broader conventions and qualities to be intelligible and effective.

The *legisign*, finally, refers to the general rules, habits and conventions that govern the structure and interpretation of prompts. Legisigns are abstract patterns that make sinsigns intelligible. They include grammatical and syntactic norms, genre expectations, and shared understandings between user and model about what constitutes a valid or meaningful input. Legisigns are not tied to any particular instance; rather, they are the structural scaffolding that enables the model to interpret and respond to prompts coherently. In the case of LLMs, legisigns encompass the conventions of natural language, the forms of prompt engineering, and the encoded patterns in the model's training data. The model recognizes certain prompt structures not because of their singular occurrence but because they belong to a system of signs that has been reinforced through repeated exposure and meaningful use. For example:

- The imperative form "Write a…" is a legisign that represents a generative task.
- The interrogative "What is…" is a legisign that cues the model to provide a definition or explanation.

These forms are recognized not because of their singular occurrence but because they belong to a system of signs encoded in the model's training data. The model has learned, through repeated exposure, that "Compare…" prompts often require contrastive analysis, while "Summarize…" prompts call for condensation of information.

Consider furthermore these prompts:

- "Explain the difference between mitosis and meiosis."
- "What are the causes of inflation in a post-industrial economy?"
- "Translate this sentence into Spanish: 'I am learning to cook.'"

Each of these prompts rely on legisigns to be intelligible. The model understands the expected response format because these prompt structures are part of a conventional system. Even more abstract legisigns include genre expectations: A prompt beginning with "Once upon a time…" invokes the narrative genre, while "In conclusion…" represents the closing of an argumentative essay.

To see how all three sign types operate simultaneously, consider the prompt:

- "Compose a eulogy for a fictional robot who sacrificed itself to save humanity."

This prompt functions as a legisign by conforming to the imperative form "Compose a…," which represents a generative task. It is a sinsign because it is a specific input entered at a particular moment. And it is rich in qualisign potential, evoking themes of sacrifice, emotion, and speculative fiction. The model responds by drawing on conventions of eulogy writing, narrative structure, and affective language – all shaped by the interplay of legisign, sinsign, and qualisign.

Another example can be:

- "Imagine you are a 19th-century philosopher. Reflect on the nature of artificial intelligence."

Here, the legisign is the imperative "Reflect on…," the sinsign is the actual prompt entered by the user, and the qualisign includes the stylistic cues of historical voice, philosophical tone, and speculative inquiry. The model's response will be shaped by all three dimensions, producing a text that simulates the voice of a 19th-century thinker while engaging with a modern concept.

What emerges from this analysis is a layered understanding of prompts as semiotic constructs. A single prompt may simultaneously function as a legisign (by conforming to linguistic or other conventions), a sinsign (as a specific input at a given time), and a qualisign (through its stylistic and affective qualities). These dimensions are not mutually exclusive but co-present, shaping the model's response in complex and interdependent ways. The presentative aspect of the prompt – how it appears, how it is structured, and how it is experienced – thus involves a dynamic interplay between possibility, actuality, and generality. Or formulated differently, we begin to recognize that prompts in LLMs are not merely functional inputs but potentially rich semiotic events that operate across different levels of meaning. By analyzing them through Peirce's typology of qualisigns, sinsigns, and legisigns, we gain insight into how prompts present themselves, how they are instantiated, and how they draw upon shared conventions to elicit intelligent responses. This Peircean framework also opens the door to further analysis using his other two sign typologies: icon, index, and symbol (which concern the representative dimension), and rheme, dicent, and argument (which concern the interpretative dimension).

**2.2.2. Prompts and the second sign typology (the representative dimension)**
The *iconic* part of prompts operates by providing a model, template, or exemplar of the desired output, allowing the LLM to recognize and replicate the essential qualities demonstrated in the example. This approach leverages the model's pattern recognition capabilities by presenting concrete instances that embody the abstract requirements of the task. This iconic strategy represents one of the most fundamental and effective approaches in contemporary prompt engineering, particularly manifesting in "few-shot" or "one-shot" scenarios where the user demonstrates the desired task through concrete examples rather than abstract instructions. The power of iconic prompting lies in its ability to communicate complex, nuanced requirements that might be difficult to articulate through purely symbolic means (Liu et al., 2023). Consider the following example:

- Input: Translate English to academic French.
  English: The problem is complex.
  French: La problématique est complexe.
  English: We need to analyze the results.
  French:

In this prompt structure, the provided translation pair serves as a multifaceted icon that communicates several layers of meaning simultaneously. The example "La problématique est complexe" functions iconically by embodying the specific qualities of the desired Interpretant: The formal register appropriate to academic discourse, the lexical sophistication expected in scholarly translation (choosing "problématique" over the more colloquial "problème"), and the structural patterns characteristic of French academic prose. The icon does not merely state the translation task abstractly but provides a concrete instantiation that the model can use as a template for generating similar outputs. The iconic function extends beyond simple input-output demonstrations to encompass various forms of qualitative specification. Style-based prompts such as "Write in the style of Ernest Hemingway" operate iconically by invoking the recognizable qualities associated with a particular author's voice - conciseness, understated emotion, dialogue-driven narrative - that the model has learned to associate with that name through its training data. Similarly, format specifications like "Provide the answer as a JSON object with keys 'author' and 'title'" function iconically by presenting a structural template that embodies the desired organizational pattern of the output. The effectiveness of iconic prompts stems from their alignment with the LLM's fundamental learning mechanism, which operates through pattern recognition and statistical association rather than rule-following (Tenney et al., 2019; Qiu et al., 2020). By providing concrete examples that instantiate abstract requirements, iconic prompts allow the model to leverage its training to identify and replicate relevant patterns without requiring explicit algorithmic instructions.

In prompting, *indexicality* serves an important function because it creates the basis for coherent and contextually grounded conversations that transcend the limitations of isolated, context-free utterances. Indexical prompts establish meaning by creating a direct connection to specific elements within the immediate communicative environment, that may consist of previous conversational turns, uploaded documents, or shared contextual information. The most direct and recognizable examples of indexical prompting involve deictic expressions – i.e., linguistic elements whose meaning depends entirely on the context of the utterance and which point to specific referents within that context (Peirce, CP 2.305). Consider these three cases:

- Input: Summarize the text above.
- Input: Elaborate on your second point.
- Input: Based on the document I just uploaded, what are the main themes?

In each instance, the indexical elements - "the text above," "your second point," "the document I just uploaded" - function as linguistic pointers that establish direct existential connections to specific Objects within the conversational or contextual environment. E.g., the phrase "the text above" creates a direct indexical link to previously presented textual material, while "your second point" establishes a connection to a specific element within the model's previous response. These indexical relationships are essential for maintaining conversational coherence and enables complex, multi-turn interactions that build upon previous exchanges. The indexical function extends beyond a simple deictic reference to encompass more sophisticated forms of contextual anchoring. Prompts that invoke shared conversational history - such as "building on our earlier discussion of semiotics" or "using the methodology you described in your previous response" – create indexical connections to distributed elements of the ongoing dialogue. These connections are important for tasks that require iterative refinement, progressive elaboration, or a synthesis of information across multiple conversational turns (Zhou et al., 2023). Without robust indexical functioning, each prompt would exist as an isolated, decontextualized utterance, severely limiting the potential for, cumulative interactions (Zamfirescu-Pereira et al., 2023). The indexical dimension of prompting thus enable the construction of extended dialogues where meaning emerge, not just from individual prompts, but from the relationships and connections established across the entire conversational sequence. This capacity for indexical reference is particularly important in complex analytical tasks such as document analysis, iterative problem-solving, and collaborative content development, where the prompt must be anchored to pre-existing informational Objects that provide the foundation for further processing.

*Symbolic* signification represents the most pervasive and fundamental mode of meaning-making in prompting, as it underpins virtually all communication conducted through natural language or formal coding systems. The symbolic dimension of prompting encompasses not only the basic linguistic elements, i.e., words, grammatical structures, syntactic patterns, but also complex discursive conventions, genre expectations, and domain-specific terminologies that have been established through collective usage and cultural transmission. Consider the apparently simple prompt: "What is the capital of Denmark?" This utterance functions symbolically at multiple levels simultaneously. Each individual word operates as a symbol whose meaning derives from established linguistic conventions: "capital" refers to a seat of government through learned association rather than natural resemblance, "Denmark" designates a specific nation-state through conventional agreement, and the interrogative structure "What is..." follows grammatical rules that have been socially established and culturally transmitted. The effectiveness of this prompt thus depends entirely on the training of the LLM, having established these symbolic relationships and their conventional applications. The symbolic function becomes even more complex when using more advanced prompting strategies, e.g., prompts that express specialized linguistic patterns developed within a user community. Take for instance prompts that begin with phrases like “Act as a senior software developer” or “Perform a SWOT analysis on the following company”. These formulaic expressions function as specialized symbols that invoke specific personas, analytical frameworks, or response patterns within the LLM. So, the phrase “Act as...” is a powerful symbolic convention that signals to the LLM that it should adopt a specific professional role or domain expertise, drawing upon training patterns associated with that identity. Similarly, analytical frameworks like “SWOT analysis,” “root cause analysis,” or “cost-benefit analysis” function as symbolic keys that activate particular structured reasoning patterns within the model. These symbols work effectively because the LLM’s training process has established statistical associations between these conventional expressions and specific types of structured, domain-appropriate responses. The symbolic dimension also encompasses more subtle linguistic conventions related to register, tone, and discourse style. Prompts that employ formal academic language signal expectations for scholarly discourse, while those using informal expressions indicate more casual interaction modes. The model’s ability to recognize and respond appropriately to these symbolic cues depends on the breadth and quality of its training data, which must encompass sufficient examples of different discourse communities and their associated conventions. Domain-specific symbolic systems present a particular challenge and opportunity in prompting. Technical fields such as programming, mathematics, or scientific research employ specialized symbolic vocabularies that carry precise meanings within their respective communities. Effective prompting in these domains requires not only familiarity with the relevant symbolic systems but also an understanding of how these symbols function within the model’s learned associations (Tenney et al., 2019).

**2.2.3. Prompts and the third sign typology (the interpretative dimension)**
When a prompt operates *rhematically*, it initiates an exploratory, divergent semiotic process that invites the LLM to navigate and sample from its learned knowledge space without constraining the response to a single correct answer or predetermined outcome. Rhematic prompts are characterized by their open-ended nature and their invitation to creative or exploratory generation rather than factual retrieval or logical deduction. They engage the model’s capacity for pattern synthesis and creative recombination, drawing upon the statistical associations learned during training to generate novel combinations of ideas, concepts, or expressions. Consider the below examples:

- “Brainstorm innovative ideas for sustainable urban transportation systems.”
- “Write a short poem about autumn in Copenhagen that captures the mood of transition.”
- “Explore the potential implications of quantum computing for information science.”
- “Generate creative metaphors for explaining artificial intelligence to non-technical audiences.”

These prompts do not seek singular, verifiable answers but instead invite exploration of a field of possibilities within the model’s learned associations. The user is essentially requesting that the LLM sample from and synthesize patterns within its training data to generate novel combinations that exhibit desired qualitative characteristics - creativity, relevance, coherence, or aesthetic appeal. The resulting Interpretant represents one possible instantiation from a vast space of potential responses, each of which could be equally valid within the parameters established by the prompt. The rhematic mode is fundamental to tasks involving creativity, ideation, conceptual exploration, and the generation of novel content that goes beyond simple information retrieval. It leverages the LLM’s capacity for pattern recognition and recombination to produce outputs that exhibit emergent properties not explicitly present in the training data but arising from the creative synthesis of learned associations. Finally, rhematic processes serve an exploratory epistemic function, helping users discover new perspectives, generate creative solutions, or explore the boundaries of conceptual domains. They are particularly valuable in early stages of research, creative projects, or problem-solving processes where the goal is to expand the space of possibilities rather than converge on specific answers.

When a prompt operates *dicently*, it engages the LLM in an informational, convergent semiotic process that seeks to establish correspondence between the model's response and external reality. Dicent prompts are characterized by their focus on factual accuracy, empirical verification, and the retrieval or confirmation of specific information that exists independently of the communicative act itself. They engage the model's capacity to access and synthesize factual knowledge encoded in its training data, producing responses that make definite claims about the world and can be subjected to verification procedures. Again consider a few examples:

- "What is the current population of Denmark according to the most recent official statistics?"
- "Confirm whether Martin Thellefsen is an Associate Professor at the University of Copenhagen."
- "List the primary components of Peirce's triadic theory of signs."
- "Identify the publication date and main findings of Shannon's foundational paper on information theory."

These prompts establish a clear expectation that the model's response should correspond to verifiable facts about external reality. The semiotic process is oriented toward information retrieval and factual confirmation rather than creative generation or logical reasoning. The resulting Interpretant is expected to function as a reliable representation of actual states of affairs that exist independently of the model's processing. The dicent mode reflects the traditional conception of information systems as repositories of factual knowledge that can be queried and retrieved through appropriate search strategies. However, in the context of LLMs, this process is mediated by the model's training data and the statistical associations learned during the training process, introducing potential issues of accuracy, currency, and bias that must be carefully considered. Dicent processes serve a confirmatory epistemic function, helping users establish factual foundations for further reasoning, verify claims, or access specific information needed for decision-making or analysis. They are essential for tasks requiring empirical grounding, fact-checking, or the establishment of reliable knowledge bases for subsequent reasoning processes. Finally, the dicent mode presents particular challenges in the context of LLM interaction, as these models may generate responses that appear factually authoritative but are actually based on statistical patterns rather than verified knowledge. Users must develop critical evaluation skills to assess the reliability of dicent responses and implement verification procedures when factual accuracy is crucial.

When a prompt operates *argumentatively*, it engages the LLM in the most high-level form of semiotic processing, where the model must not merely retrieve information or generate creative content but follow systematic reasoning procedures to reach logically warranted conclusions. Argumentative prompts are characterized by their provision of logical frameworks, reasoning procedures, or systematic methodologies that constrain and guide the model's processing toward structured, reasoned outputs. They transform the LLM from a simple information retrieval system or creative generator into a reasoning engine capable of following complex logical procedures and applying systematic analytical frameworks. More examples:

- “Using chain-of-thought reasoning, solve this multi-step mathematical problem: [problem statement with explicit reasoning steps]”
- “Apply the principles of domain analysis as outlined by Hjørland to evaluate the following list of terms for their relevance to information science, providing systematic justification for each assessment.”
- “Given the premises that (1) all effective communication requires shared understanding and (2) LLMs lack genuine understanding, construct a logical argument about the limitations of human-AI communication.”
- “Conduct a systematic comparative analysis of Peircean and Saussurean semiotics, organizing your reasoning according to the following analytical framework: [detailed framework specification]”

These prompts establish complex logical structures that guide the model through systematic reasoning processes. Rather than simply requesting information or creative output, they provide methodological frameworks, logical premises, or reasoning procedures that the model must follow to generate structured, warranted conclusions. The resulting Interpretant represents not just information or creative content but the outcome of a systematic reasoning process that can be evaluated according to logical criteria such as validity, consistency, and coherence. The argumentative mode represents the most advanced application of prompt engineering, requiring users to understand not only the model’s capabilities but also the logical structures and reasoning procedures appropriate to their analytical goals. It enables the construction of complex tasks that leverage the model’s pattern recognition capabilities while constraining them within systematic logical frameworks. Argumentative processes serve a generative epistemic function, enabling users to construct new knowledge through systematic reasoning rather than simply retrieving existing information or generating creative content. They are essential for complex analytical tasks such as theoretical analysis, systematic comparison, logical argumentation, and the application of methodological frameworks to novel problems.

Taken together, then, these nine types of signs – involved in three semiotic dimensions, the presentative, representative, and interpretative – offer a comprehensive semiotic framework for analyzing prompts. In short, they allow us to dissect the prompt’s internal structure, its referential orientation, and its interpretative affordances. Or formulated differently, the theoretical and methodological value of this semiotic framework lies in its capacity to reveal the prompt as a layered, relational, and interpretative sign. It moves us beyond surface-level analysis and into a deeper understanding of how prompts are potentially meaning-making. Moreover, this framework is not merely descriptive; it is generative, and most importantly, perhaps, it prepares the ground for theorizing about prompts as dynamic communicative phenomena, situated within a triadic relation between user, model, and context. Something which we, in the previous pages, only have taken for granted or at best alluded to. Let us therefore now address the prompt and prompting from the perspective of the “Dynacom model”.

**3. Prompts as a process of communication**

The analytical transition from the prompt as semiotic entity to communicative process is justified by the inherent dynamism of Peirce’s sign relation. The interpretant is not a static endpoint but a generative moment that can itself become a new representamen. In prompting, each response can serve as a new prompt, creating a recursive communicative loop. The Dynacom captures this recursive and dialogic nature, emphasizing the temporal and interactive dimensions of prompting. When we analyze prompts as signs, we identify their formal conditions: The representamen (the prompt itself), the object (the intended referent or domain), and the interpretant (the response or understanding). However, this triadic relation alone does not guarantee communication. For prompting to become communicative, the interpretant must be

significant – that is, it must resonate within a shared context and produce a meaningful effect. This is where dynacom enters: It theorizes the conditions under which the interpretant becomes a cominterpretant, a shared understanding that enables dialogue, learning, or generative output.

The DynaCom model posits that communication is successful when it produces a significance-effect – a communicational outcome in which the interpreter grasps the intended meaning of the utterer. This effect is not automatic; it depends on three formal conditions: (1) communication must occur within a shared *universe of discourse*, (2) the utterer and interpreter must possess *collateral experience*, and (3) the *cominterpretant* must be activated (Thellefsen, Sørensen & Thellefsen 2011). These conditions resonate strongly with Peirce's semiotic categories and offer a pragmatic framework for analyzing prompting as a communicational act.

In prompting, the *universe of discourse* corresponds to the shared context between the user and the AI system. This includes not only the linguistic conventions of the prompt but also the epistemic domain it invokes – whether it be scientific, poetic, legal, or colloquial and so on. A prompt like "Explain quantum entanglement in simple terms" presupposes a discourse shaped by scientific knowledge and pedagogical intent. Without this shared universe, the AI's response may drift into irrelevance or incoherence.

The second condition, *collateral experience*, refers to the background knowledge and interpretive habits that both parties bring to the communicational event (Sørensen, Thellefsen & Thellefsen, 2014). In human-AI prompting, this is asymmetrical: The user brings lived (phenomenological) experience, cultural nuance, and situational awareness and so on, while the AI system brings statistical associations derived from training data. Nevertheless, successful prompting hinges on the alignment of these experiences. For instance, when a user prompts "Write a haiku about autumn," the AI must draw on its learned understanding of poetic form and seasonal imagery to produce a response that resonates with the user's expectations. This alignment is further shaped by the user's own evolving mental model of the LLM, i.e., collateral experience is not static but is continuously updated through interaction and refining strategy for subsequent prompts.

The third condition, the *cominterpretant*, is perhaps the most Peircean of the three. It refers to the interpretive act that confirms the intended meaning has been grasped. In prompting, this is reflected in the user's evaluation of the AI's output – whether the response is deemed relevant, insightful, or emotionally resonant.

The cominterpretant is not static; it evolves through iterative prompting, feedback, and refinement. It is here that Peirce's final interpretant finds its analogue: the habitual understanding that emerges over time through repeated communicational success.

Prompting, when modeled via the Dynacom framework, emerges not as a static input-output mechanism but as a communicative process. This process unfolds across multiple interpretative stages, each involving the transformation of signs and interpretants, and each shaped by the interplay of Peirce's nine sign types – qualisign, sinsign, legisign; icon, index, symbol; rheme, dicent, argument – as well as the triadic interpretant structure: intentional, effectual, and cominterpretant. The Dynacom model provides a scaffold for understanding how meaning is constructed, negotiated, and refined in human-AI interaction in the context of prompting large language models.

The process begins with **Stage 1**: Initial intentionality and sign formation. Here, the *utterer* – the human user – experiences a pre-linguistic intentional state, a mental interpretant (I_user), which arises from the recognition of a gap between their current knowledge and a desired understanding. This gap motivates the communicative act. The user's intention, though not yet verbalized, is already semiotically active. It is shaped by qualisigns, such as the mood or tone the user wishes to convey; by legisigns, such as the general linguistic conventions they plan to employ; and by rhemes, which represent the potential meanings they hope to explore. The transformation of this internal state into a prompt involves encoding these abstract intentions into an intentional interpretant that can be processed by the model. This encoding draws on symbolic elements (linguistic structures), indexical cues (contextual references), and iconic forms (examples or analogies that resemble the desired output). This stage also presupposes a shared *universe of discourse* – a conceptual domain within which both the utterer and the interpreter (the LLM) operate. The universe of discourse includes the language, topics, and genres that are mutually intelligible to both parties. For example, a prompt about quantum physics assumes that both the user and the model have access to relevant terminology and conceptual frameworks. The *collateral experience* of the user – their background knowledge, expectations, and familiarity with the model's capabilities – also plays a crucial role in shaping the prompt. It informs how the user formulates their request, what they assume the model can understand, and how they anticipate the model will respond.

**Stage 2** marks the emergence of the prompt as the primary representamen ($S_1$). This prompt is now a sinsign – a specific, actualized linguistic event situated in time and space. It is also a legisign insofar as it conforms to grammatical, syntactic and semantic norms, and it carries qualisign features that influence its expressive force. For example, the prompt "Write a love letter to the moon" operates symbolically through its imperative structure, indexically by referencing the moon as a known cultural object, and iconically by resembling the genre of a love letter. It also functions as a rheme, suggesting a possibility rather than asserting a fact. In contrast, a prompt like "Explain the gravitational effects of the moon on Earth's tides" is more dicent in nature, asserting a factual inquiry and guiding the model toward an argumentative or explanatory response. These distinctions illustrate how prompts are multi-modal signs that represent the user's intention in a form designed to elicit a specific kind of interpretative behavior from the model.

**Stage 3** involves machine interpretation which also concerns the preparation for the "later" response generation. Upon receiving the prompt, the LLM functions as the interpreter of the communication process. It processes the representamen through its trained parameters, generating an internal computational state – its effectual interpretant – which reflects the actual effect the prompt has on the model's system. This includes the activation of relevant knowledge patterns, probabilistic associations, and structural generation mechanisms. Furthermore, the communicational interpretant also begins to "take its form"; that is, there a process of determination concerning what the communication is about, and seen from the perspective of the LLM this refer to, inter alia, shared conventions and expectations (with the utterer, the human user) that will guide the model's subsequent behavior – qua its training data, linguistic norms, genre knowledge etc.

Figure 1 seen below depicts the most relevant semiotic elements of Stages 1-3 in the process of prompting understood as communication.

## Figure 1

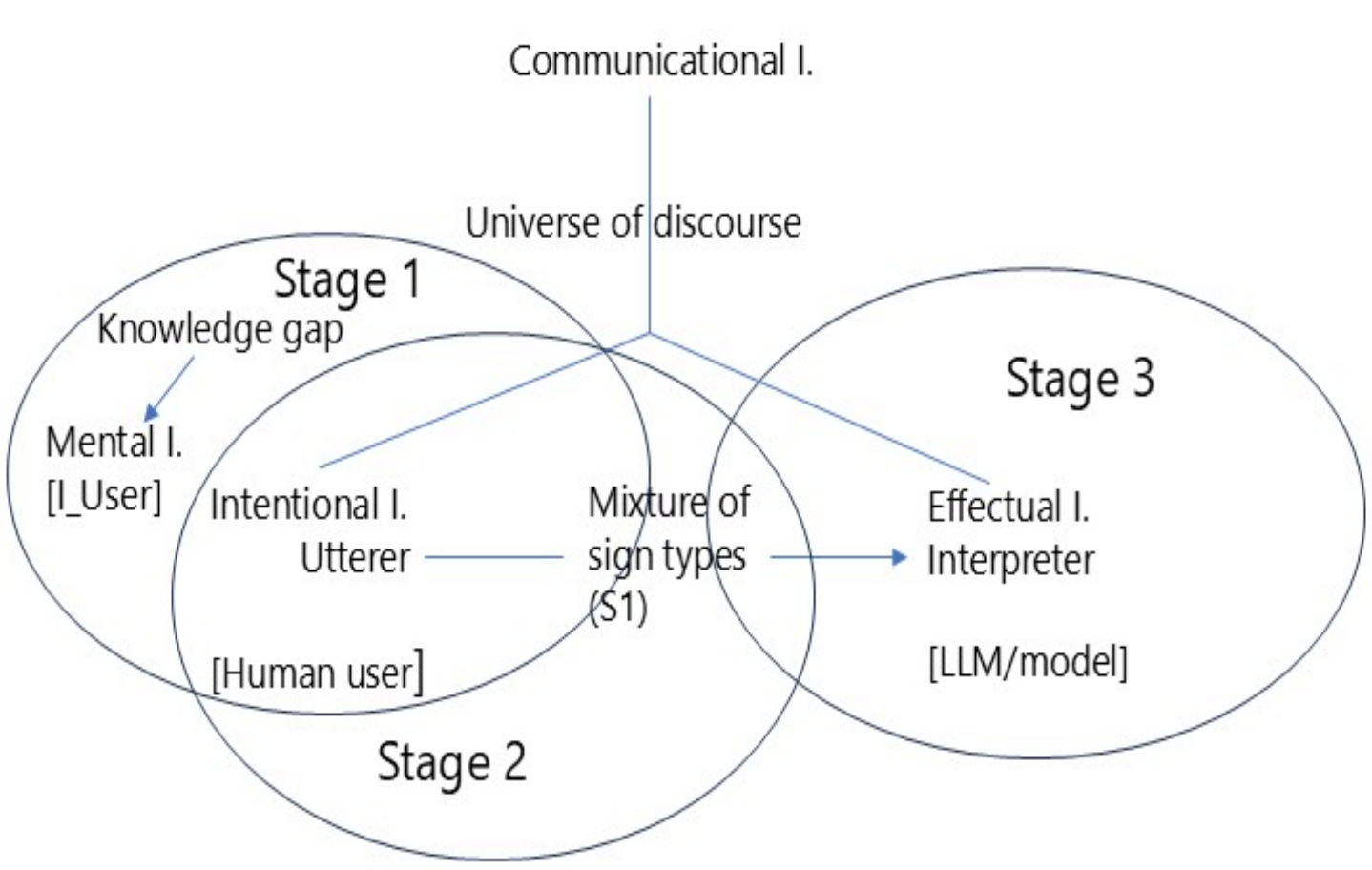

**Figure 1**: Depicting the process of prompting as communication involving the semiotic Stages 1-3.

**Stage 4** now sees the model externalize its interpretation as a textual output: Involving a new intentional interpretant (in the communication process), which expresses the LMM's inferred understanding of the user's communicative goal. The model's response is externalized as a textual output ($S_2$), which becomes a new sinsign – an actualized linguistic event – and simultaneously a new representamen for the human user now having the role as interpreter. This transformation exemplifies the principle of semiotic continuity: Every interpretant can become a new representamen in the next interpretative act. The user now engages with the model's response, evaluating its adequacy, accuracy, and relevance. This evaluation constitutes a new mental interpretant ($I_2$), shaped by the user's *intentional interpretant* (their original goal), the *effectual interpretant* (the actual impact of the model's response), and the *cominterpretant* (the shared understanding between user and model).

Figure 2 tries to capture how the LLM now has the role of utterer and the human user the role of interpreter respectively; thus, the LLM has the (semiotic) capacity to produce a string of sign types endowing the process with an intentional interpretant relative to a universe of discourse and the initial representamina of the human user.

Figure 2

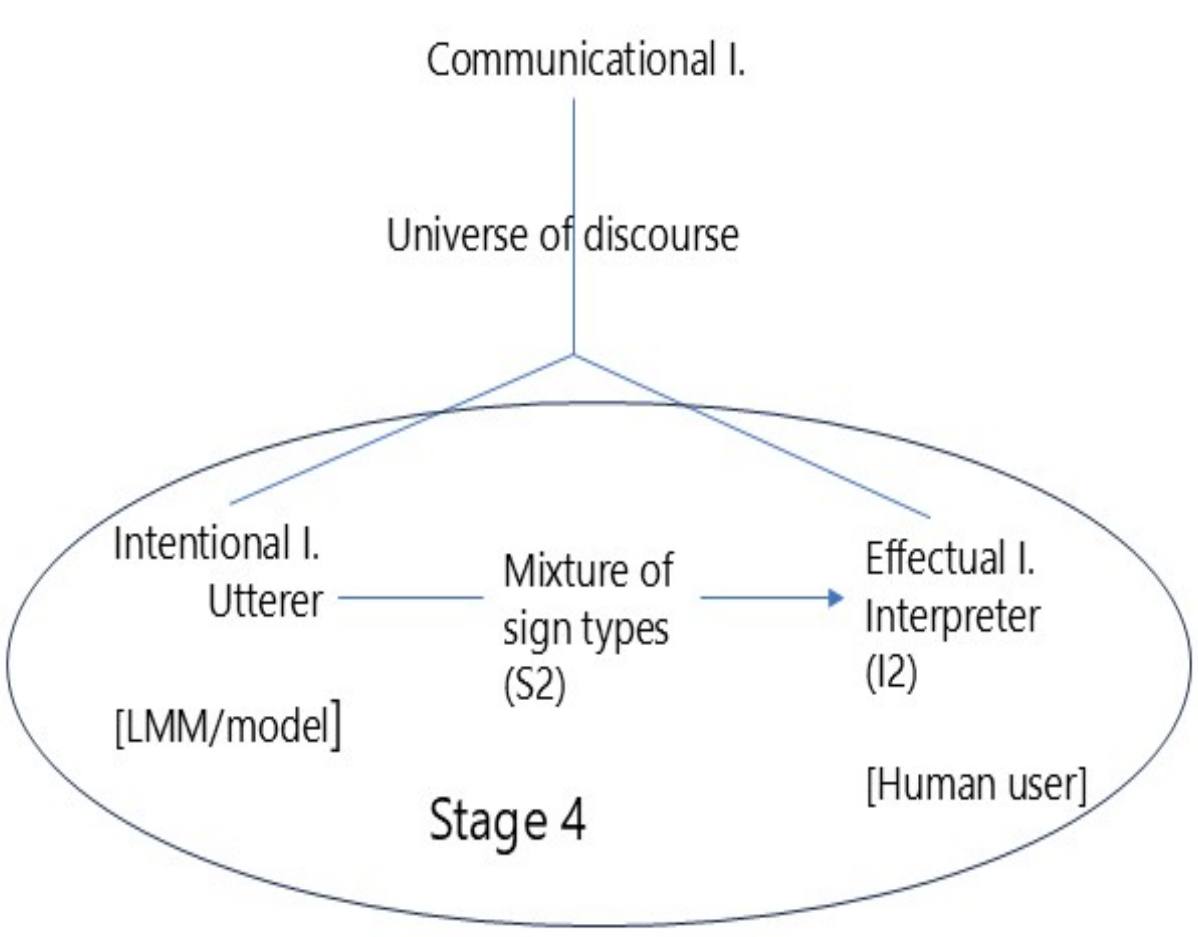

**Figure 2:** The LLM as utterer and the human user as interpreter; the LLM communicates a string of sign types (as a result of the prompt) with the purpose of evoking communicational interpretants; that is, there are established interpretants in the interpreter which are relevant for "closing" his/her knowledge gap.

If the human user does not (yet), however, experience a stable, satisfactory interpretation of the prompt that resolves the initial gap between their knowledge state and desired understanding, Stages 5 and 6 may follow.

**Stage 5** thus involves user re-interpretation and chain continuation. The user formulates a new prompt ($S_3$) that builds upon the previous exchange. This prompt may refine the original intention, correct misunderstandings, or introduce new contextual elements. For example, if the initial prompt was "Describe the feeling of walking through a foggy forest at dawn" and the model responded with a generic description, the user might follow up with "Make it more poetic and include sensory details." This new prompt is again a sinsign, shaped by legisign conventions and enriched with qualisign cues. It may also function as an argument, asserting a specific directive for stylistic enhancement. The user, as utterer, continues to draw upon their collateral experience and the shared universe of discourse to guide the communicative process.

**Stage 6** represents iterative refinement and convergence. The semiotic chain—$S_1 \rightarrow I_1/S_2 \rightarrow I_2/S_3 \rightarrow I_3/S_4 \rightarrow$ …—continues until the user reaches a final communicational interpretant: A stable interpretation that resolves the initial gap between their knowledge state and desired understanding. This endpoint is not merely the result of a single prompt but the culmination of a dynamic communicative process involving multiple sign transformations, interpretative acts, and feedback loops. Each turn in the dialogue contributes to the accumulation of contextual meaning, and each prompt-response pair functions as a semiotic unit within a broader communicative event. In short, we can also understand this process of prompting as the continuous representation of the dynamic object through a series of immediate objects – revealing more and more aspects of the dynamic object.

Throughout this process, all nine Peircean sign types are active. Qualisigns convey characters, e.g., shaping the expressive tone of prompts and responses. Sinsigns instantiate these signs in specific linguistic events. Legisigns provide the structural rules that govern interpretation. Icons resemble desired outputs or genres. Indexes point to contextual referents or situational cues. Symbols operate through conventional linguistic meanings. Rhemes suggest possibilities. Dicents assert facts or inquiries. Arguments guide reasoning and structure complex responses. The interpretants – intentional, effectual, and cominterpretant – mediate the relationship between signs and meaning, ensuring that the communicative process remains responsive, adaptive, and goal-oriented.

In sum, prompting is not a mere act of inputting text into a machine. It is a semiotic dialogue – a communicative process that unfolds through iterative sign (type) formation, interpretation, and refinement; also making this process be open to semiotic analysis concerning information, knowledge and emotion (Sørensen, Thellefsen & Thellefsen, 2016). The Dynacom model points towards the complexity of the process of prompting as communication, revealing how meaning is constructed through the interplay of signs and interpretants. Effective prompting thus requires not only linguistic competence but also semiotic awareness: An understanding of how signs function, how meaning evolves, and how human-machine interaction can be strategically managed to achieve communicative goals. The concepts of utterer and interpreter, collateral experience, and universe of discourse are essential to this model, as they frame the conditions under which signs are produced, interpreted, and transformed. By attending to these semiotic layers, we can better understand how prompts function, how they are experienced, and how they elicit meaningful responses from intelligent systems.

## 4. Some implications of the semiotic perspective on prompts and prompting for information science and knowledge organization

Building on the semiotic framework developed through Peirce's triadic model, his nine sign types, and the dynacom model of communication, we can now consider some broader implications for information science and knowledge organization. At the center of these considerations is the large language model, which, when viewed through a semiotic lens, emerges not merely as a computational tool but as a novel semiotic resource – one that fundamentally challenges the traditional boundaries between human and machine interpretation.

In classical information science, knowledge organization has been grounded in human-centric systems: classification schemes, controlled vocabularies, metadata standards, and ontologies. These systems presume a stable relationship between signs and meanings, mediated by human intentionality and interpretative authority. Furthermore, the human subject, as utterer, is assumed to possess the epistemic agency to produce meaningful signs, while the system functions as a passive repository or retrieval mechanism. Meaning-making, in this paradigm, is anchored in human cognition and stabilized through institutionalized structures of representation.

The LLM, however, operates in different ways. It does not retrieve pre-existing knowledge; it generates interpretants dynamically in response to prompts. This generative capacity positions the LLM as a semiotic agent capable of producing signs that are meaningful within specific universes of discourse, even though it lacks consciousness or intentionality.

From a Peircean perspective, each prompt issued to an LLM initiates a sign relation: The prompt is the representamen, the object is the conceptual or referential domain it evokes, and the interpretant is the output generated by the model. Crucially, this interpretant is not fixed – it is shaped by the model's training data, algorithmic structure, and the context of the prompt. The interpretant of the prompt becomes meaningful when it resonates with the user's collateral experience and aligns with a shared universe of discourse. This dynamic interpretative process, as captured by the Dynacom model, reveals that prompting is a communicative act, and the LLM is a participant in that communication.

The involvement of Peirce's nine sign types further illuminates the LLM's semiotic complexity. The model's outputs may instantiate legisigns (general linguistic or other conventions), sinsigns (specific textual instances), and qualisigns (stylistic or affective qualities). They may function iconically (mimicking human discourse), indexically (responding to user input), and symbolically (drawing on cultural codes). Moreover, they may take the form of rhemes (possibilities), dicents (assertions), or arguments (reasoned claims). This typological richness suggests that LLMs operate across all dimensions of Peirce's semiotic schema, making them unprecedented semiotic resources.

This reconceptualization has potential significant theoretical and analytical implications for knowledge organization and information seeking, especially when prompting is foregrounded as the central communicative mechanism. Four implications for knowledge organization seem to stand out:

Firstly, prompting challenges the assumption that meaning and meaning-making are exclusively human domains. If LLMs can produce interpretants that are coherent, relevant, and contextually appropriate in response to prompts, then meaning-making becomes a distributed process – shared between human and machine. The prompt, as a semiotic act, encodes the user's intentional interpretant and initiates a communicative exchange in which the model contributes its own interpretative labor. This calls for a redefinition of epistemic authority in information systems: The LLM, though lacking consciousness, functions as an interpreter capable of generating signs that contribute to knowledge construction. Prompting thus becomes a site of epistemic negotiation, where human and machine co-construct meaning.

Secondly, prompting destabilizes the boundary between representation and interpretation. Traditional knowledge organization systems are designed to represent knowledge through fixed categories, descriptors, and taxonomies. LLMs, by contrast, interpret prompts and generate new representations dynamically. Each prompt is a representamen that evokes a specific object, and the model's output is an interpretant that may reframe, recontextualize, or even reconceptualize that object. This shift from static representation to dynamic interpretation requires new theoretical models that can account for the fluidity, contextuality, and dialogical nature of meaning. Prompting becomes a form of real-time knowledge negotiation rather than mere retrieval, and knowledge organization must adapt to this interpretative dynamism.

Thirdly, prompting invites a reconfiguration of metadata and indexing practices. If meaning is generated dynamically and contextually through prompts, then metadata must be flexible, adaptive, and capable of capturing interpretative dimensions. This may involve designing metadata schemas that reflect semiotic features – such as sign types, discourse universes, and collateral experience – rather than merely descriptive attributes. Prompts themselves may become metadata: They reveal user intention, contextual framing, and interpretative expectations. Indexing systems must therefore be capable of accommodating the semiotic richness of prompts and the interpretative variability of model responses.

Fourthly, prompting offers a richer foundation for understanding the epistemological status of AI-generated knowledge. Rather than evaluating prompt outputs solely in terms of truth or accuracy, we can assess them in terms of their semiotic coherence, communicative efficacy, and alignment with shared interpretative frameworks. A prompt such as "Explain the concept of justice in Plato's Republic" does not merely seek factual retrieval; it initiates a semiotic exchange in which the model must interpret genre, philosophical context, and user intention. The resulting output can be evaluated not only for correctness but for its resonance with the user's intentional interpretant and its capacity to engage a cominterpretant. Prompting thus becomes a method for epistemic inquiry, and the model's responses are situated within a broader semiotic ecology.

In relation to information seeking, the query and the specific query techniques, i.e., knowing how to use specific codes, descriptors and combinations hereof e.g., by means of Boolean operators, have traditionally been key for searching information. The concepts of control and verifiability have consistently played a central role in the practice of information retrieval. However, the very nature of searching information changes because the traditional query is replaced by the prompt. As we have demonstrated, the prompt is a much richer semiotic artifact. Thus, in relation to information seeking we have selected four areas that seems to be imperative. Firstly, where the traditional query formulation predominantly is a symbolic activity, and users translate needs into keywords, prompting is a more sophisticated process of sign construction (as semiosis). An effective prompt is a hybrid sign that leverages not only symbols (descriptors, language), but also icons (examples, styles) and indices (contextual references), which allows the user to shift from merely stating a need to actively demonstrating it, thus enabling the user to express their information goal in more nuanced ways. Secondly, the process of searching for information is no longer a single search-and-retrieve action but an iterative, conversational dialogue, modelled as a semiotic chain. Or formulated differently, final interpretants are realized through iterative prompting in the information search process – actualizing parts of a (vast) meaning potential (immediate interpretants) through series of dynamic interpretants. This dynamic process aligns with established models of information behavior such as Kuhlthau's ISP model (Kuhltau 1991). Initial open-ended rhematic prompts mirror the exploratory stages of seeking where users feel uncertainty. As the interaction progresses, the user can employ more specific dicent prompts for factual clarification, which is analogue to the critical "focus" stage in the ISP model. The argumentative prompt, furthermore, allows for a sophisticated synthesis of information moving beyond simple retrieval. Thirdly, the search engine becomes a semiotic partner simulating a dialogue where meaning is co-created. Interaction is reframed as an epistemic negotiation where knowledge is co-constructed; and the concept of relevance is transformed from match of retrieved documents to communicative success in the generated responses. Fourthly, and finally, information literacy is redefined as semiotic competence. "Prompt engineering" can be considered a critical new literacy that is fundamentally semiotic. This new competence is the ability to effectively construct signs to elicit the desired interpretants. Such an approach will move beyond instrumentalism toward a semiotic ethics of prompting – one that respects the complexity of meaning(-making) and the continuity of interpretation, entailing a user's responsibility for the signs they construct, a critical awareness of the societal biases encoded in the model's semiotic system, and a commitment to verifying the interpretants it generates. And it invites analysis of prompts for their affective and conceptual affordances (as immediate interpretants), their responses (as dynamic interpretants), not just for correctness but also for interpretive depth and variation, as well as tracking usage patterns (final interpretants) to understand how prompts evolve into conventions or norms. "Prompt engineering", therefore, in short, involves the skill of blending iconic, indexical and symbolic elements and strategically shifting between rhematic, dicent and argumentative modes to successfully navigate the LLM's vast knowledge base. In conclusion, then, positioning the LLM as a semiotic resource reframes its role in information science and knowledge organization and information seeking. It is not merely a tool for processing language but a participant in meaning-making through prompts and prompting – as a generator of interpretants (effectual, intentional, and cominterpretant) within dynamic communicative contexts. Prompts are not neutral inputs; on the contrary they are semiotic acts that represent intention, invoke context, and initiate interpretation. This perspective challenges traditional boundaries between human and machine interpretation; furthermore, it calls for new models of epistemic agency; and, finally, it invites a reimagining of the theoretical and methodological foundations of knowledge organization in the age of computational semiosis. By attending to the semiotic structure of prompts and the interpretative behavior of LLMs, we can begin to develop a more nuanced, responsive, and dialogical approach to organizing, accessing, and evaluating knowledge in digital environments.

**5. A few concluding thoughts**

Throughout the previous pages, we have tried to reconceptualize prompts and prompting not as mere technical inputs but as genuine semiotic acts embedded within dynamic communicative processes. Drawing on Peirce's triadic model of sign (representamen, object, interpretant), his nine sign types (qualisign, sinsign, legisign; icon, index, symbol; rheme, dicent, argument), and the Dynacom model of communication, we have aimed to show how prompting involves iterative meaning-making between human and machine. Each prompt initiates a semiotic chain (of potential stages), shaped by the user's intentional interpretant, the model's effectual interpretant, and the shared cominterpretant that enables mutual intelligibility.

Prompts represent tone, genre, and intention (qualisign), instantiate specific linguistic events (sinsign), and conform to general conventions (legisign). They operate representationally through relations of similarity (icon), contextual reference (index), and cultural convention (symbol), and interpretatively as possibilities (rheme), assertions (dicent), or reasoned claims (argument). The LLM, though not conscious, therefore functions as a semiotic interpreter – generating outputs as prompts that participate in meaning-making within shared universes of discourse and shaped by the user's collateral experience.

This semiotic reframing has potentially important implications for information science and knowledge organization. It namely challenges the human exclusivity of meaning-making; furthermore, it destabilizes static models of representation and thus calls for adaptive metadata practices; this also invites a more context-sensitive approach to evaluating AI-generated knowledge potential such as prompts. In short, prompting is not just a technical skill – it is a practice of semiotic and epistemic stance that redefines how knowledge is organized, interpreted, and co-constructed in the age of intelligent systems.